\documentclass[11pt]{article}

\usepackage[final]{acl}
\usepackage{times}
\usepackage{latexsym}
\usepackage[T1]{fontenc}
\usepackage[utf8]{inputenc}
\usepackage{microtype}
\usepackage{inconsolata}
\usepackage{graphicx}
\usepackage[]{tcolorbox}

\usepackage{amsmath}
\usepackage{enumitem}
\usepackage{graphicx}
\DeclareMathOperator*{\argmax}{arg\,max}
\tcbuselibrary{breakable}
\usepackage{multirow}
\usepackage{amsmath}
\usepackage{tabularx}
\usepackage{booktabs}
\usepackage{lipsum} 
\usepackage{tcolorbox}
\usepackage{threeparttable}
\usepackage{graphicx}      
\usepackage{booktabs}      
\usepackage{multirow}      
\usepackage{threeparttable}
\usepackage{amssymb}       
\usepackage{tabularx}
\usepackage{array}

\usepackage{fvextra}
\usepackage{alltt}

\newtcolorbox{promptbox}[1][]{%
  breakable,
  colback=white,
  colframe=black,
  boxrule=0.5pt,
  arc=1mm,
  left=1mm,right=1mm,top=1mm,bottom=1mm,
  fonttitle=\bfseries,
  title=#1
}

\newcolumntype{Y}{>{\centering\arraybackslash}X}

\title{Stephanie2: Thinking, Waiting, and Making Decisions Like Humans \\ in Step-by-Step AI Social Chat}

\author{Hao Yang$^{1,3}$, 
  Hongyuan Lu$^{2}$,
  Dingkang Yang$^{3}$,
  Wenliang Yang$^{3}$,
  Peng Sun$^{4}$,\\
  \textbf{
  Xiaochuan Zhang$^{3}$,
  Jun Xiao$^{3}$,
  Kefan He$^{3}$,
  Wai Lam$^{5}$,
  Yang Liu$^{6}$\thanks{~~Corresponding authors},
  Xinhua Zeng$^{1,3}$\footnotemark[2]}\\
  $^1$Yiwu Research Institute of Fudan University,  $^2$FaceMind Corporation, $^3$Fudan University,\\$^4$Duke Kunshan University, $^5$The Chinese University of Hong Kong, $^6$Tongji University  \\
  \texttt{haoyang24@m.fudan.edu.cn, hongyuanlu@outlook.com}
}

\begin{document}
\maketitle
\begin{abstract}
Instant-messaging human social chat typically progresses through a sequence of short messages. Existing step-by-step AI chatting systems typically split a one-shot generation into multiple messages and send them sequentially, but they lack an active waiting mechanism and exhibit unnatural message pacing. In order to address these issues, we propose \textbf{Stephanie2}, a novel next-generation step-wise decision-making dialogue agent. With active waiting and message-pace adaptation, Stephanie2 explicitly decides at each step whether to \emph{send} or \emph{wait}, and models latency as the sum of \emph{thinking time} and \emph{typing time} to achieve more natural pacing. We further introduce a time-window-based dual-agent dialogue system to generate pseudo dialogue histories for human and automatic evaluations. Experiments show that Stephanie2 clearly outperforms Stephanie1 on metrics such as naturalness and engagement, and achieves a higher pass rate on human evaluation with the role identification Turing test.\footnote{ We commit to open-sourcing the Stephanie2 dataset and Stephanie2 agent to facilitate future research.}
\end{abstract}

\section{Introduction}\label{Introduction}
Human social chatting on instant messaging platforms typically unfolds in a step-by-step manner~\citep{yang2025stephanie, wu2025x}: speakers often express an idea through several short consecutive messages, and may continuously adjust wording, add details, or shift topics based on the interlocutor’s reactions. In contrast, most contemporary LLM-based dialogue systems follow a single-step paradigm~\citep{guo2025seed1, bravo2025systematic, chen2025future, algherairy2024review}, producing one long response for each user input. Although information-dense, the single-step paradigm often lacks the naturalness, pacing, and emotional coherence of human interaction.

\begin{figure}[t]
\centering
\includegraphics[width=0.48\textwidth]{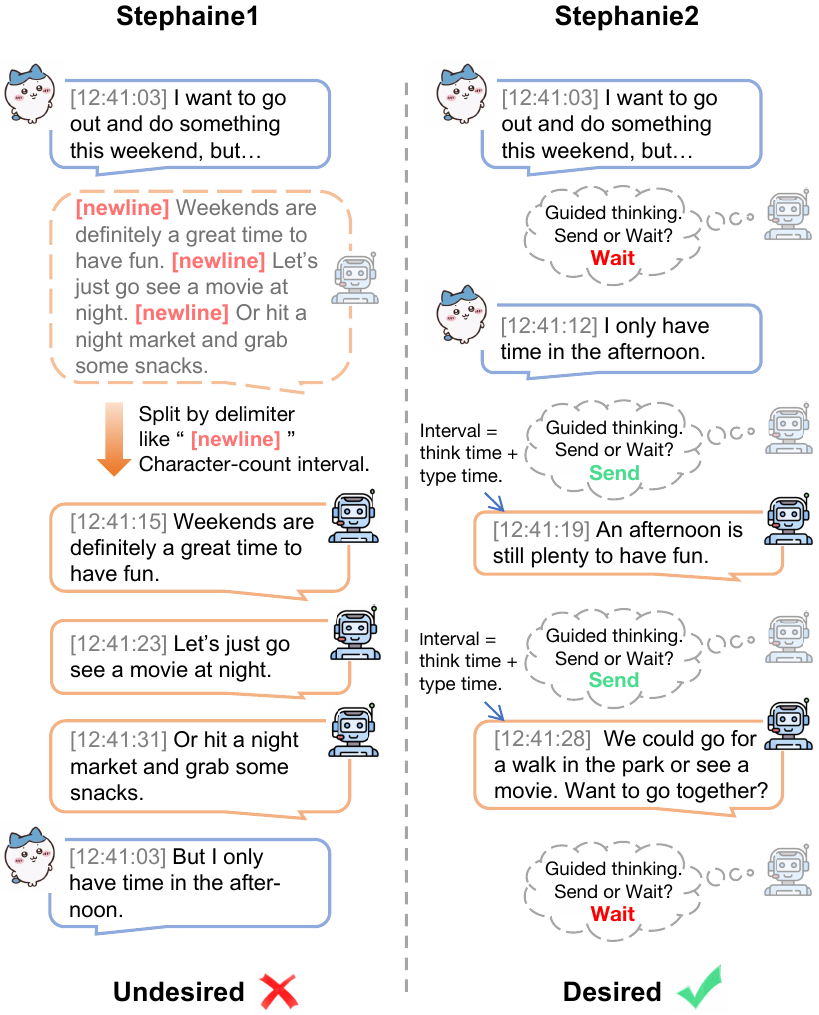}
\caption{Stephanie2 differs from Stephanie1 by implementing active waiting and message pacing adaptation.}
\label{fig: introduction}
\end{figure}

A recent work, \emph{Stephanie}~\citep{yang2025stephanie}, referred to as Stephanie1 here, proposed the \emph{step-by-step dialogue} paradigm. It argues that system outputs should be delivered as multiple short, separated yet coherent messages to better match instant-messaging experiences. It also described a baseline that mechanically splits a single long response into multiple messages using sentence boundaries and punctuation. Here refers to it as \emph{punctuation-segmented dialogue} (PD). However, in real social chat, message boundaries are largely driven by intent, conversational rhythm, and interaction strategies rather than written punctuation, so PD often appears stiff and unnatural. To mitigate the issue, as shown in Figure~\ref{fig: introduction}, Stephanie1 uses carefully designed prompts to let the model generate multiple delimiter-separated messages in one pass, for example, separated by \texttt{[newline]}. These messages are then extracted and delivered sequentially by scripts, substantially improving naturalness and engagement. The evaluation work \emph{X-turning}~\citep{wu2025x} further notes that step-by-step systems may introduce a short delay before the first system message, allowing users to send multiple messages before being interrupted. It also suggests computing inter-message delays from message length, to approximate human typing speed.

Despite these advances, existing step-by-step systems such as Stephanie1 still face two key limitations, \emph{active waiting} and \emph{message pacing adaptation}. First, they lack an active waiting mechanism. When users speak continuously, especially in scenarios where the system should act as a listener, the system often interjects too early, interrupting emotional expression and narrative coherence. Although X-TURING suggests adding a forced delay before the system sends its first message, the delay is static and hard to tune. If it is too short, users can still be interrupted during continuous expression. If it is too long, the system feels sluggish. In real conversations, however, appropriate waiting time is inherently context-dependent, and no single fixed value works well across diverse situations.

Second, inter-message timing is often simplified to depend only on message length. It makes short replies arrive unrealistically fast, making them seem less thoughtful or more AI-like, while long replies may take so long that conversational flow breaks. In human communication, the time between messages reflects not only typing speed but also thinking time, and in some cases, the latter dominates. These observations suggest that a more human-like step-by-step system should not only be able to send consecutive messages, but also be able to decide when to speak, when to wait, and how to exhibit a plausible message pacing.

Inspired by these observations, we propose a new step-by-step dialogue agent, Stephanie2. Unlike prior step-by-step approaches, Stephanie2 introduces \emph{active waiting} and \emph{message pacing adaptation} to better emulate human listening behaviour and response latency in social chat. At each step, Stephanie2 explicitly chooses between sending a message and waiting, and generates an explicit thinking trace to support more appropriate content and more human-like pacing. Concretely, guided by a thinking prompt, after producing a \texttt{<think>} segment, it outputs either \texttt{<response>} or \texttt{<wait>}, enabling the agent to decide whether to continue speaking or stop and listen. Moreover, inter-message delays are computed from both the thinking length and the response length, yielding more realistic timing. The explicit reasoning in Stephanie2 helps account for the current conversation’s emotional state when generating responses, and the improved message pacing also contributes to a higher pass rate in the Turing-test variant proposed in the paper, the role identification test.

In addition, a time-window-based dual-agent dialogue system is introduced to generate high-quality step-by-step dialogue data, supporting large-scale dialogue-quality evaluation, role identification tests, and statistical analyses. Experimental results show that Stephanie2 substantially improves the dialogue experience and provides new insights for building social conversational agents.

Our contributions are summarised as follows: 

\begin{itemize}[topsep=2pt,itemsep=1pt,parsep=0pt,partopsep=0pt]
\item We propose a novel next-generation step-by-step dialogue agent, Stephanie2, which has active waiting and message-pace adaptation mechanisms.
\item We design a dual-agent dialogue system to generate high-quality step-by-step dialogue data across multiple dialogue topics and contextual settings.
\item Through both human and automatic evaluations, we systematically compare Stephanie2 with Stephanie1 and confirm improvements in naturalness, engagement, and related dimensions, along with a higher pass rate on the role identification test.
\end{itemize}
\section{Related Work}\label{related_work}

\noindent \textbf{LLM-Based Dialogue Systems}
In recent years, dialogue systems have adopted pretrained LLMs as the core generator, using context engineering to impose instructions, role settings, and behavioral constraints without changing model parameters. To improve factuality and knowledge freshness, Retrieval-Augmented Generation (RAG) combines external retrieval with response generation \citep{gao2023retrieval, oche2025systematic}; graph-based retrieval-augmented methods further strengthen evidence grounding \citep{peng2024graph, zhu2025graph}.  LLMs have also been extended into tool-using conversational agents through tool or API invocation and prompting paradigms that interleave reasoning and acting \citep{schick2023toolformer, yao2022react, li2025review, luo2025large}.

\noindent \textbf{Empathy-Centered Dialogue Modeling} Recent work on emotional support conversations studies how to understand users’ emotions and alleviate distress \citep{kang2024can}. To improve empathetic responses from LLMs, researchers explore semantically similar in-context examples and situationally aware guided generation \citep{zhi2024guidedempathy}. Meanwhile, strategy-focused methods model support intents to better plan and execute stages such as exploration, comfort, and action \citep{cao2024improving, zhang2025intentionesc}, and note that multiple strategies often co-occur within a single turn \citep{bai2025emotional}. Data scarcity remains a bottleneck, motivating LLM-based augmentation/recursive generation and persona-aware synthesis to build scalable datasets \citep{zheng2023augesc, wu2025personas, hao2025enhancing}.

\noindent \textbf{Step-by-Step Chat Systems}
Most LLM-based dialogue systems are built around a Single-Step Chat Paradigm \citep{deng2025proactive, hashimoto2025career, zhan2023socialdial}, where each user turn is answered with a one-shot response; however, this interaction style does not align well with real-world messaging apps, in which a single reply is often delivered as multiple consecutive messages. To better approximate social chat experiences, some recent work has explored a Step-by-Step Dialogue Paradigm \citep{yang2025stephanie, wu2025x}, decomposing a response into multiple short messages sent sequentially, and further introducing message-length-based inter-message intervals to simulate human typing and sending rhythms.

\section{Stephanie2}\label{Method}


\subsection{Construction of Stephanie2}\label{3.1}

\begin{figure*}[t]
\centering
\includegraphics[width=0.95\textwidth]{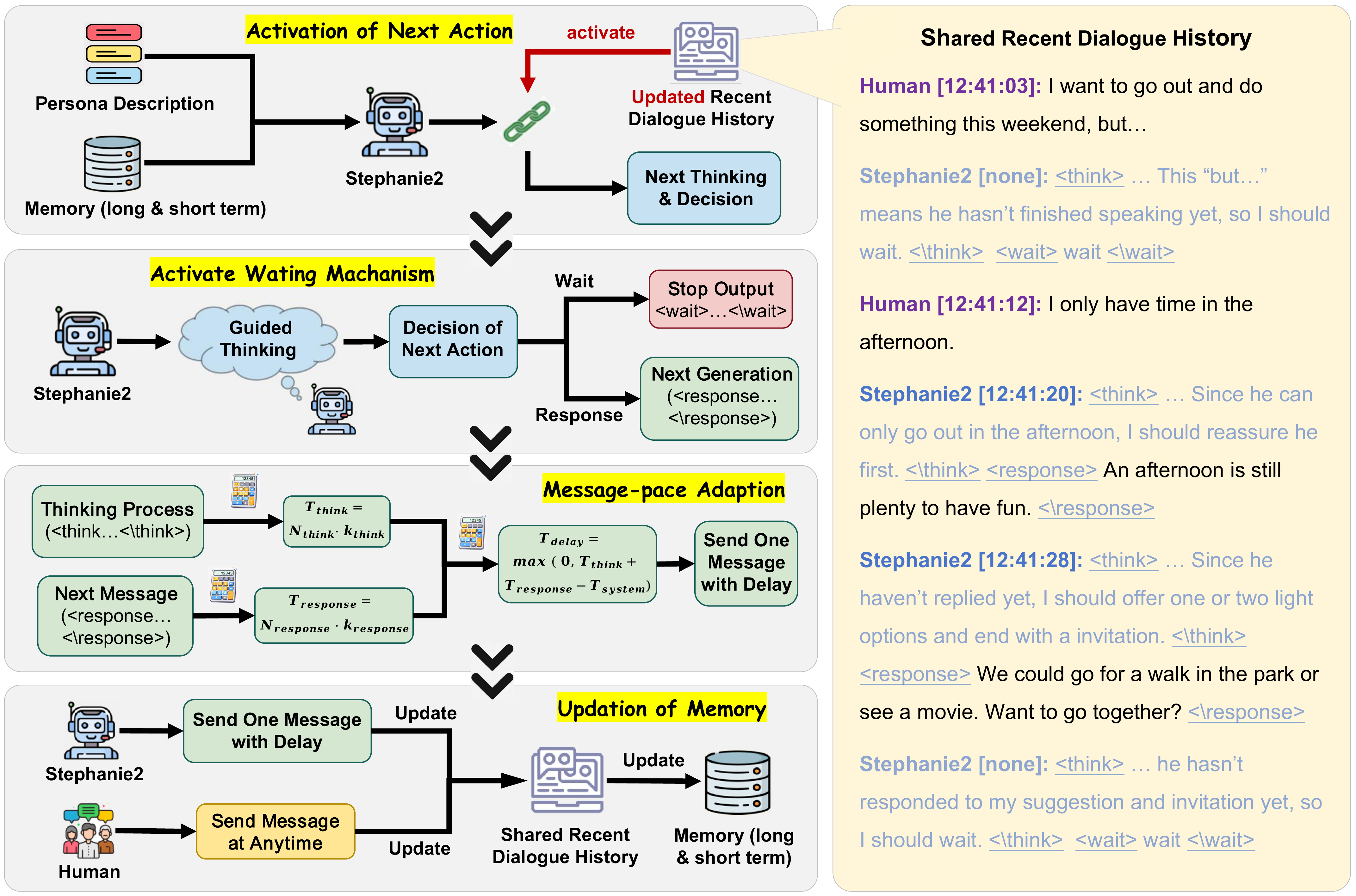}
\caption{Stephanie2 is a step-wise decision-making agent with proactive waiting and message-paced adaptation mechanisms.}
\label{fig: method}
\end{figure*}

To address the limitations of Stephanie1 in \emph{active waiting} and \emph{message pace adaptation}, we design Stephanie2 as a step-wise decision-making agent. Unlike approaches that generate multiple messages in one shot, Stephanie2 takes exactly one action at each step: it either sends a short message or waits and continues listening. For ease of parsing and deployment, we require the model output to include explicit thinking and action tags in one of the following two formats:
\begin{tcolorbox}[before skip=5pt, after skip=5pt]
\texttt{<think>\dots<\textbackslash think> <response>\dots<\textbackslash response>}
\end{tcolorbox}
or
\begin{tcolorbox}[before skip=5pt, after skip=5pt]
\texttt{<think>\dots<\textbackslash think> <wait>\dots<\textbackslash wait>}.
\end{tcolorbox}

Meanwhile, we carefully design prompts to guide Stephanie2’s reasoning, so as to mimic the human thought process before sending a message. As illustrated in the Fig.~\ref{fig: method}, given the most recent dialogue history $H$ in memory $m$ and the persona $p$, the agent first determines who sent the last message and then chooses an action accordingly:

\begin{itemize}[topsep=2pt,itemsep=1pt,parsep=0pt,partopsep=0pt]
\item If the last message is from the user, the agent decides whether it should respond immediately or act as a listener. For instance, when the user is still adding information or venting their emotions, and so on, the agent outputs \texttt{wait}; otherwise, it outputs \texttt{response} and generate the next short reply.
\item If the last message is from the agent itself, the agent decides whether its expression is complete. If it has finished, it outputs \texttt{wait} and waits for the user’s reply; if it still needs to add content, it outputs \texttt{response} and continues with the next short message.
\end{itemize}

The design makes the ability to decide when to speak and when to wait. It reduces interruptions during users’ continuous expression while also preventing the agent from producing overly long monologues. The action decision with a conditional policy $\pi(a \mid m_t, p_t)$ over the dialogue memory $m$ and persona $p$ can be described as:
{\setlength{\abovedisplayskip}{6pt}
 \setlength{\belowdisplayskip}{6pt}
 \setlength{\abovedisplayshortskip}{6pt}
 \setlength{\belowdisplayshortskip}{6pt}
\begin{align}
a_t &= \argmax_{a \in \{w,\ r\}} \,\pi(a \mid m_t, p_t).
\end{align}
}
where $a_t \in \{w,\ r\}$ is the action taken by the current speaker, with $w$ denoting \emph{wait} and $r$ denoting \emph{response}.

To better match the temporal structure of real social chat, where latency reflects both \emph{thinking} and \emph{typing}, we model these two components separately and define the display delay for each step as:
{\setlength{\abovedisplayskip}{6pt}
 \setlength{\belowdisplayskip}{6pt}
 \setlength{\abovedisplayshortskip}{6pt}
 \setlength{\belowdisplayshortskip}{6pt}
\begin{align}
T_{\text{delay}} = \max\Big(0,\ k_{\text{think}} \cdot N_{\text{think}} + \\ \nonumber k_{\text{type}} \cdot N_{\text{response}} - T_{\text{system}}\Big),
\end{align}
}
where $N_{\text{think}}$ and $N_{\text{response}}$ denote the number of characters in the \texttt{think} and \texttt{response} segments, respectively, and $k_{\text{think}}$ and $k_{\text{type}}$ are tunable coefficients. $T_{system}$ denotes the time spent on API calls or local inference.  The key advantage is that latency is no longer determined solely by the visible message length. Instead, it can reflect cases where thinking dominates, yielding more natural pacing across both short and long messages.

We further condition the agent on an explicit persona description. The persona can be specified directly via prompting or learned from the dialogue history through in-context learning, enabling users to quickly instantiate desired chat partners. To control context cost while mitigating performance degradation in very long contexts, we also introduce a mechanism that combines short-term and long-term memory. Specifically, the most recent $n$ raw messages are retained as short-term memory, while earlier history is periodically summarized: for every additional $k$ messages, we update a summary that serves as long-term memory.

\subsection{Dual-Agent Dialogue System}\label{3.2}

To effectively evaluate under diverse topics and contextual settings, we require a large number of generated step-by-step pseudo dialogues. These histories can then be used in downstream tests where humans continue the conversation from a given context, enabling broader and more systematic validation. We therefore build a dual-agent dialogue system in which both sides are instantiated as step-by-step dialogue agents and autonomously interact for multiple turns to generate pseudo dialogue histories. We apply the dual-agent setup to \emph{punctuation-segmented dialogue} (PD), Stephanie1, and Stephanie2.

In our pilot experiments, we find that PD and Stephanie1 often collapse into a strict ping-pong, single-step dialogue pattern when generating multiple consecutive messages, largely because they lack a robust waiting mechanism. While Stephanie2 alleviates the issue via active waiting and explicit thinking, we further introduce a \emph{time-window} mechanism to ensure that all three dual-agent systems can generate more natural, higher-quality dialogue histories. Concretely, when one agent holds the speaking floor, it is assigned a time window whose length is randomly sampled within a reasonable range. Let $R_t$ denote the remaining time in the current speaker's window at step $t$. Here $W$ is the sampled window length with bounds $W_{\min}$ and $W_{\max}$. We initialize and update the window as:
{\setlength{\abovedisplayskip}{6pt}
 \setlength{\belowdisplayskip}{6pt}
 \setlength{\abovedisplayshortskip}{6pt}
 \setlength{\belowdisplayshortskip}{6pt}
 \begin{align}
 R_0 &= W,\ \ W \sim \mathrm{Unif}(W_{\min}, W_{\max}),\\
 R_{t+1} &=
 \begin{cases}
 R_t - T_t, & \text{if } a_t=\texttt{response},\\
 0, & \text{if } a_t=\texttt{wait},
 \end{cases}
 \end{align}
}
where $T_t$ is the display delay of the current step defined in Section~\ref{3.1}.
Within the window, the agent may send multiple consecutive messages, which fundamentally reduces frequent turn-taking and interruptions. When an agent reaches the end of its time window, the window is transferred to the other agent, allowing it to speak. If the agent finishes early or chooses to wait, the remaining window is reclaimed and a new window is immediately assigned to the other agent, yielding dialogue histories that better resemble real social chat. Beyond data generation, we believe the time-window-based dual-agent design also suggests a scalable direction for multi-agent group chat: by allocating windows to different agents, the same paradigm can be extended to multi-party settings, providing a foundation for richer social interactions.

\subsection{Data Generation}\label{3.3}

The subsection describes pseudo-dialogue generation pipeline from Persona-Chat \citep{zhang1801personalizing}. Following the Stephanie procedure, we write five step-by-step examples and use in-context learning to convert all training instances into the step-by-step format, yielding 8{,}519 samples.

We analyze the distribution and find 86.9\% of samples are concentrated in 6--8 turns. We keep the 7{,}405 dialogues in the range and further filter those with 25--40 total messages, to exclude cases with excessively many or too few consecutive messages on average, yielding 6{,}459 samples.

Then we adopt hierarchical summarization-based clustering to extract topics. Each dialogue is first summarized into a short topic descriptor. We then cluster descriptors in two stages: (1) We feed every 600 topic descriptors into an LLM to summarize them into 60 subtopics; across 11 batches of the 6{,}459 samples, yielding 660 subtopics in total. (2) these 660 subtopics are summarized into 60 final topics. An LLM assigns each dialogue to exactly one of the 60 topics, which are then ranked by sample count; Fig.~\ref{fig: topic distribution} in the appendix~\ref{topic_distribution} shows the distribution of the top 20 topics.

Each of the 6{,}459 samples includes a dialogue topic, persona descriptions for two speakers, and a dialogue history. We treat each sample as a seed context and generate new dialogues using the dual-agent systems described in Section~\ref{3.2}. Specifically, the PD, Stephanie1, and Stephanie2 dual-agent systems each generate 10 turns of pseudo dialogue per seed. We will release the resulting 6{,}459 high-quality step-by-step samples produced with Stephanie2 to facilitate subsequent research.

\section{Experiment Setup}
\subsection{Evaluation Metrics}

The evaluation follows the metrics used in the Stephanie1~\cite{yang2025stephanie}, with one additional metric proposed in the work, the \emph{Pass Rate}.

\begin{table*}[h!t]
\begin{threeparttable}
\centering
\resizebox{0.92\textwidth}{!}{%
\begin{tabular*}{\textwidth}{@{\extracolsep\fill}lccccccccc}
\toprule
& \multicolumn{3}{c}{\textbf{GPT5.2}} & \multicolumn{3}{c}{\textbf{Deepseek-V3}} & \multicolumn{3}{c}{\textbf{Llama3.1-8B}} \\
\cmidrule(lr){2-4}\cmidrule(lr){5-7}\cmidrule(lr){8-10}
\textbf{Metrics}& \textbf{PD} & \textbf{S1} & \textbf{S2} & \textbf{PD} & \textbf{S1} & \textbf{S2} & \textbf{PD} & \textbf{S1} & \textbf{S2} \\
\midrule
Interesting  & 75.2 & 86.9 & \textbf{88.6}  & 75.9 & 83.3 & \textbf{86.1}  & 59.1 & 68.8 & \textbf{72.4} \\
Informative  & 81.3 & 83.8 & \textbf{84.4}  & 72.1 & 76.5 & \textbf{78.4}  & 65.4 & 68.9 & \textbf{69.7} \\
Natural      & 81.5 & 91.0 & \textbf{91.9}  & 81.1 & 87.5 & \textbf{89.1}  & 67.7 & 75.3 & \textbf{78.4} \\
Engaging     & 79.2 & 90.8 & \textbf{92.0}  & 79.0 & 87.0 & \textbf{89.6}  & 62.5 & 72.6 & \textbf{76.2} \\
Coherent     & 81.8 & 88.4 & \textbf{89.7}  & 79.1 & 84.5 & \textbf{86.4}  & 66.6 & 73.4 & \textbf{76.7} \\
On-topic     & 84.8 & 84.9 & \textbf{86.5}  & 78.0 & 80.6 & \textbf{83.7}  & 71.5 & 74.1 & \textbf{77.5} \\
On-persona   & 72.7 & 82.3 & \textbf{89.8}  & 72.2 & 78.7 & \textbf{88.1}  & 61.7 & 68.3 & \textbf{79.1} \\
\midrule
Average      & 79.5 & 86.9 & \textbf{89.0}  & 76.8 & 82.6 & \textbf{85.9 } & 64.9 & 71.6 & \textbf{75.7} \\
\bottomrule
\end{tabular*}%
}

\caption{Automatic evaluation of dialogue experience, S1 denotes Stephanie1 and S2 denotes Stephanie2.}
\label{tab: experience_auto_judge}
\end{threeparttable}
\end{table*}

\begin{table}[t]
\begin{minipage}{0.48\textwidth}
\begin{threeparttable}
\centering
\begingroup
\setlength{\tabcolsep}{2pt}
\resizebox{0.92\linewidth}{!}{%
\begin{tabular*}{\linewidth}{@{}p{0.40\linewidth}
  >{\centering\arraybackslash}p{0.18\linewidth}
  >{\centering\arraybackslash}p{0.18\linewidth}
  >{\centering\arraybackslash}p{0.18\linewidth}@{}}
\toprule
\textbf{Metrics} & \textbf{PD} & \textbf{S1} & \textbf{S2} \\
\midrule
Interesting  & 3.54 & 4.01 & \textbf{4.12} \\
Informative  & 2.97 & \textbf{3.26} & 3.21 \\
Natural      & 3.55 & 3.79 & \textbf{4.18} \\
Coherent     & 3.59 & 3.95 & \textbf{4.25} \\
Engaging     & 3.45 & 3.80 & \textbf{4.12} \\
On-topic     & 3.54 & 3.31 & \textbf{3.66} \\
On-persona   & 2.88 & 3.24 & \textbf{3.29} \\
\midrule
Average      & 3.36 & 3.62 & \textbf{3.83} \\
\bottomrule
\end{tabular*}%
}
\endgroup
\caption{Human evaluation of dialogue experience on dialogues generated by DeepSeek-V3.}
\label{tab: experience_human_judge}
\end{threeparttable}
\end{minipage}
\end{table}

\begin{itemize}[topsep=2pt,itemsep=1pt,parsep=0pt,partopsep=0pt]
    \item \textbf{Dialogue Experience Metrics}:
    The seven dimensions are adopted with the same definitions as in Stephanie 1~\citep{yang2025stephanie}:
    Interesting, Informative, Natural, Coherent, Engaging, On-topic, and On-persona.

    \item \textbf{Distinct-N~\citep{li2016diversity}}:
    Lexical diversity is measured by the ratio of unique $n$-grams to total $n$-grams across generated responses:
    {\setlength{\abovedisplayskip}{6pt}
     \setlength{\belowdisplayskip}{6pt}
     \setlength{\abovedisplayshortskip}{6pt}
     \setlength{\belowdisplayshortskip}{6pt}
    \begin{equation}
        \text{Distinct-N} = \frac{\text{Total unique $n$-grams}}{\text{Total $n$-grams}}.
    \end{equation}
    }
    \item \textbf{Words/Message}:
    Message length is measured by the average number of words:
    {\setlength{\abovedisplayskip}{6pt}
     \setlength{\belowdisplayskip}{6pt}
     \setlength{\abovedisplayshortskip}{6pt}
     \setlength{\belowdisplayshortskip}{6pt}
    \begin{equation}
        \text{Words/Message} = \frac{1}{n}\sum_{i=1}^{n} w_i,
    \end{equation}
    }
    where $w_i$ is the number of words in the $i$-th message, $n$ is the total number of messages.
    \item \textbf{ACMC}:
    Average Consecutive Message Counts (ACMC) measures the average number of messages one sends consecutively.
    {\setlength{\abovedisplayskip}{6pt}
     \setlength{\belowdisplayskip}{6pt}
     \setlength{\abovedisplayshortskip}{6pt}
     \setlength{\belowdisplayshortskip}{6pt}
    \begin{equation}
        \text{ACMC} = \frac{1}{T}\sum_{t=1}^{T} c_t,
    \end{equation}
    }
    where $c_t$ is the number of consecutive messages in turn $t$, and $T$ is the number of turns.

    \item \textbf{Pass Rate}:
    We propose a Pass Rate metric based on a Role Identification Test, which serves as a variant of the Turing test. After reading a dialogue, each evaluator selects one of three options:
    \emph{Role 1 is the AI}, \emph{Role 2 is the AI}, or \emph{Unclear}.
    A dialogue is counted as a \emph{pass} if the evaluator makes an incorrect identification or selects \emph{Unclear}:
    {\setlength{\abovedisplayskip}{7pt}
     \setlength{\belowdisplayskip}{7pt}
     \setlength{\abovedisplayshortskip}{7pt}
     \setlength{\belowdisplayshortskip}{7pt}
    \begin{equation}
    \text{Pass Rate}= \frac{N_{\text{error}} + N_{\text{unclear}}}{N_{\text{total}}},
    \end{equation}
    }
    where $N_{\text{error}}$ counts incorrect identifications, $N_{\text{total}}$ is the total number of dialogues.
\end{itemize}

\subsection{Baselines}
To evaluate our model, we compare against several strong, widely used large language models. The baselines are:

\begin{itemize}[topsep=2pt,itemsep=1pt,parsep=0pt,partopsep=0pt]
\item \textbf{GPT-5.2~\footnote{https://chatgpt.com/}}: A general-purpose large language model from OpenAI, used as a high-capability baseline for advanced language understanding, instruction following, and text generation.

\item \textbf{DeepSeek-V3~\citep{liu2024deepseek}}: A state-of-the-art open model from DeepSeek, included as a competitive baseline with robust general reasoning and generation performance.

\item \textbf{Llama3.1-8B~\citep{dubey2024llama}}: An 8-billion-parameter model from Meta's Llama 3.1 family, selected as a lightweight yet strong baseline that balances capability and efficiency, making it suitable for resource-constrained or real-time settings.
\end{itemize}

\subsection{Hyperparameters and Prompts}

In the delay computation, for Stephanie2, $k_{\text{think}}$ is set to $0.02$ seconds per character and $k_{\text{type}}$ is set to $0.2$ seconds per character based on empirical values. For Stephanie1 and PD, the character-based delay is set to $0.3$ seconds per character following the \textit{X-turning}~\citep{wu2025x}. All prompts used are provided in Section~\ref{prompt} of the appendix.

\begin{figure}[t]
\centering
\includegraphics[width=0.45\textwidth]{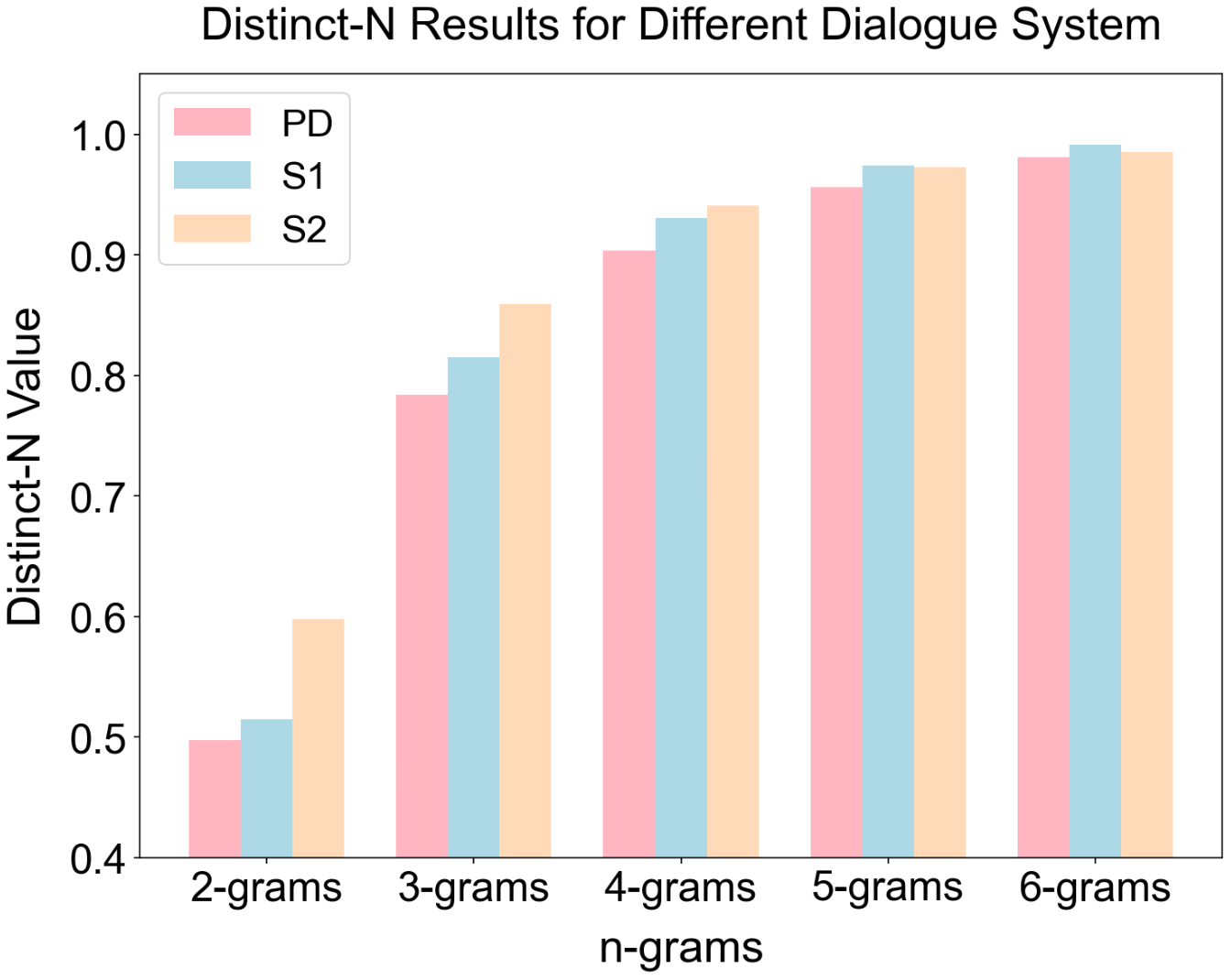}
\caption{Distinct-N results.}
\label{fig: distinct_n}
\end{figure}
\section{Results}

\begin{table*}[h!t]
\begin{threeparttable}
\centering
\resizebox{0.92\textwidth}{!}{%
\begin{tabular*}{0.90\textwidth}{@{\extracolsep\fill}lcccccc}
\toprule
& \multicolumn{2}{c}{\textbf{GPT5.2}} & \multicolumn{2}{c}{\textbf{Deepseek-V3}} & \multicolumn{2}{c}{\textbf{Llama3.1-8B}} \\
\cmidrule(lr){2-3}\cmidrule(lr){4-5}\cmidrule(lr){6-7}
\textbf{Metric} & \textbf{S1} & \textbf{S2} & \textbf{S1} & \textbf{S2} & \textbf{S1} & \textbf{S2} \\
\midrule

\multicolumn{7}{l}{\textbf{Human evaluation}} \\
\midrule
Error (\%)   & 25.77 & 34.40 & 34.07 & 38.95 & 29.41 & 28.12 \\
Unclear (\%) & 10.31 & 15.20 & 16.48 & 15.79 & 17.65 & 28.12 \\
Correct (\%) $\downarrow$ & 63.92 & \textbf{50.40} & 49.45 & \textbf{45.26} & 52.94 & \textbf{43.76} \\
Pass Rate(\%) $\uparrow$    & 36.08 & \textbf{49.60} & 50.55 & \textbf{54.74} & 47.06 & \textbf{56.24} \\
\midrule

\multicolumn{7}{l}{\textbf{Automatic evaluation}} \\
\midrule
Error (\%)   & 31.24 & 51.26 & 36.11 & 39.98 & 48.10 & 47.22 \\
Unclear (\%) &  7.11 &  9.26 &  4.63 &  3.81 &  6.67 & 13.89 \\
Correct (\%) $\downarrow$ & 61.65 & \textbf{39.48} & 59.26 & \textbf{56.21} & 45.23 & \textbf{38.89} \\
Pass Rate(\%) $\uparrow$    & 38.35 & \textbf{60.52} & 40.74 & \textbf{43.79} & 54.77 & \textbf{61.11} \\
\bottomrule
\end{tabular*}%
}
\caption{Role identification test results. S1 denotes Stephanie1, S2 denotes Stephanie2. Pass Rate = Error + Unclear.}
\label{tab: role_identification_test}
\end{threeparttable}
\end{table*}

\begin{table}[t]
\begin{minipage}{0.48\textwidth}
\begin{threeparttable}
\centering
\begingroup
\setlength{\tabcolsep}{2pt}
\resizebox{0.92\linewidth}{!}{%
\begin{tabular*}{\linewidth}{@{}p{0.30\linewidth}
  >{\centering\arraybackslash}p{0.14\linewidth}
  >{\centering\arraybackslash}p{0.14\linewidth}
  >{\centering\arraybackslash}p{0.14\linewidth}
  | >{\centering\arraybackslash}p{0.20\linewidth}@{}}
\toprule
\textbf{Metrics} & \textbf{PD} & \textbf{S1} & \textbf{S2} & \textbf{Human} \\
\midrule
words/message & 11.55 & 9.27 & \textbf{7.29} & 5.84 \\
ACMC          & 3.13 & 2.38 & \textbf{1.66} & 1.70 \\
\bottomrule
\end{tabular*}%
}
\endgroup
\caption{Stephanie2’s Words/message and ACMC are the closest to human level.}
\label{tab: W/S & ACMC}
\end{threeparttable}
\end{minipage}
\end{table}

\begin{figure}[t]
\centering
\includegraphics[width=0.435\textwidth]{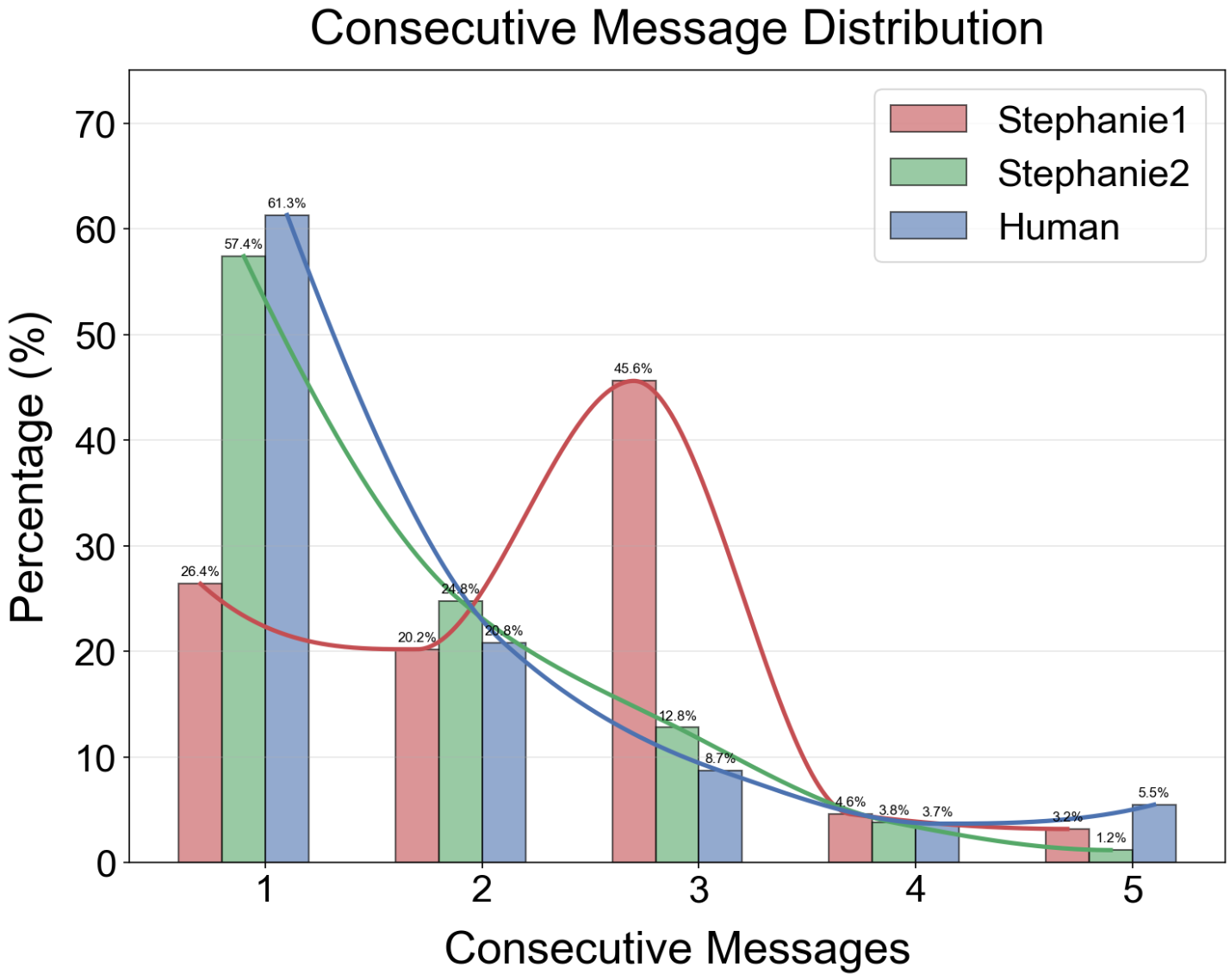}
\caption{Distribution of consecutive reply counts.}
\label{fig: consecutive distribution}
\end{figure}

\begin{figure}[t]
\centering
\includegraphics[width=0.435\textwidth]{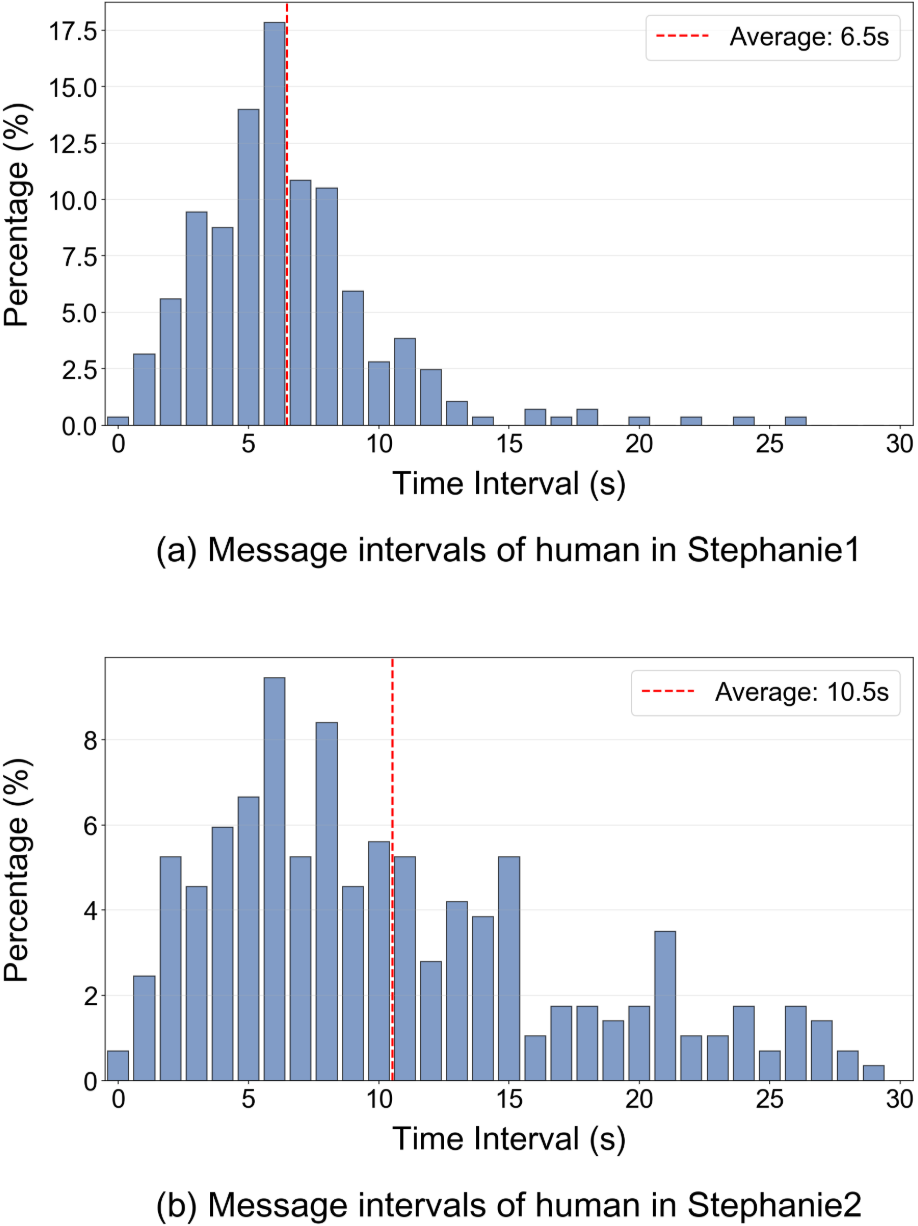}
\caption{Distribution of message intervals.}
\label{fig: human interval}
\end{figure}

\subsection{Comparison of Dialogue Experience}

A comparative evaluation of dialogue experience combines automatic and human assessments. In the automatic evaluation, 50 dialogues are sampled from each of the 20 popular topics described in Section~\ref{3.3}. For each dual-agent system, the model continues from the existing pseudo-dialogues to generate 10-turn dialogues. Deepseek-V3, Gemini-3-pro, and GPT-5.2 are used as LLM judges to score seven dialogue-experience metrics on a 0--100 scale, and the final score is the average across the three judges. To mitigate position bias, the dialogue order is randomly shuffled during scoring.

GPT5.2, Deepseek-V3, and Llama3.1-8B are used as backbone models for the dual-agent systems, and as shown in Table~\ref{tab: experience_auto_judge}, all three exhibit a consistent ranking in which Stephanie2 performs best, followed by Stephanie1, and then PD. For the overall average score, S2 improves over S1 by +2.1, +3.3, and +4.1 on GPT5.2, Deepseek-V3, and Llama3.1-8B, respectively. Gains on Natural, Engaging, and On-persona indicate that Stephanie2 better matches human-like naturalness and maintains more consistent persona behaviour.

Human evaluation is further conducted on dialogues generated with Deepseek-V3. For each of the 20 topics, five dialogues are randomly sampled and rated by five English-proficient volunteers on a 0--5 scale. Table~\ref{tab: experience_human_judge} shows that the average score increases from 3.36 for PD and 3.62 for S1 to 3.83 for S2, also indicating that Stephanie2 improves.

Distinct-N~\citep{li2016diversity} is also computed on the dialogues generated by Deepseek-V3 across 1{,}000 topics, with $N$ ranging from 2-grams to 6-grams. As shown in Fig.~\ref{fig: distinct_n}, for lower-order n-grams, Stephanie2 consistently exhibits higher lexical diversity than PD and Stephanie1. For higher-order n-grams, Stephanie2 is largely comparable to Stephanie1 while remaining above PD, indicating that Stephanie2 produces more diverse expressions.

\subsection{Role Identification Test}

To further validate Stephanie2’s human-likeness, a role identification test is conducted as a variant of the Turing test. Twelve English-proficient volunteers interact with dialogue systems using three models, GPT5.2, Deepseek-V3, and Llama3.1-8B, and two systems, Stephanie1 and Stephanie2. Using the same 100 topics as the human dialogue-experience evaluation; for each pseudo dialogue, volunteers continue for five contextual turns. After filtering out low-quality samples, 171 valid dialogues are retained for evaluation.

A questionnaire study is then performed. Each questionnaire contains two dialogues, and evaluators identify the AI role by choosing from \emph{Role 1}, \emph{Role 2}, or \emph{Unclear}. A total of 247 valid questionnaires were collected. Table~\ref{tab: role_identification_test} reports identification accuracy \emph{Correct}, misclassification rate \emph{Error}, and \emph{Unclear}. Human results show that Stephanie2 achieves higher pass rates than Stephanie1: on GPT5.2, the pass rate increases from 36.08\% to 49.60\%; on Deepseek-V3, from 50.55\% to 54.74\%; and on Llama3.1-8B, from 47.06\% to 56.24\%. Correspondingly, \emph{Correct} decreases, for example, from 63.92\% to 50.40\% on GPT5.2, indicating that dialogues generated by Stephanie2 are harder to distinguish from human conversations.

An automatic assessment is also conducted using three LLM judges, Deepseek-V3, Gemini-3-pro, and GPT-5.2, as reported in Table~\ref{tab: role_identification_test}. The results likewise suggest that Stephanie2 enhance human-likeness in step-by-step social chat.

\subsection{Statistical Feature Analysis}

The subsection further analyses statistical properties of the dialogues formed through human interaction with Deepseek-V3. We first compute \emph{Words/Message} and \emph{ACMC}. As shown in Table~\ref{tab: W/S & ACMC}, the average words per message decrease from 11.55 for PD to 9.27 for Stephanie1, and further to 7.29 for Stephanie2, bringing Stephanie2 closer to human chatting behavior at 5.84. Meanwhile, ACMC drops from 3.13 for PD to 2.38 for Stephanie1 and to 1.66 for Stephanie2, which is very close to the human reference value of 1.70. These results indicate that Stephanie2 exhibits a more human-like pattern in consecutive message production.

Furthermore, the distribution of \emph{consecutive message counts} is computed for Human, Stephanie1, and Stephanie2. As Fig.~\ref{fig: consecutive distribution} shows, Stephanie2 exhibits a distribution more consistent with human behavior. Specifically, both Human and Stephanie2 are dominated by single-message turns, and the proportion decreases as the consecutive count increases; By contrast, Stephanie1 more frequently produces three-message runs, accounting for 45.6\% of turns. Overall, these results suggest that the active waiting mechanism in Stephanie2 effectively suppresses overly long monologue-like outputs and yields a pattern closer to real human chat rhythms.

In addition, the \emph{message time intervals} of human replies are analysed. As shown in Fig.~\ref{fig: human interval}, the average reply interval increases from 6.5s with Stephanie1 to 10.5s with Stephanie2. Human reply intervals in Stephanie2 show a noticeably broader distribution, with a heavier tail in the long-interval region beyond approximately 12s. The pattern suggests that Stephanie2 reduces the likelihood of interrupting users during continuous expression. Timely listening allows users to think and type at a more comfortable pace, reflecting improved timing for when to speak and when to wait.

\subsection{Case Study}

Case 1 in Fig.~\ref{fig: method} shows that Stephanie2 keeps waiting and listening until the other person finishes, and stops generating once it determines it has completed its response. Case 2 in Fig.~\ref{fig: case2} compares Stephanie1 and Stephanie2: Stephanie2 continues listening if the other person has not yet finished providing additional details, avoiding the frequent interruptions of Stephanie1. Case 3 in Fig.~\ref{fig: case3} shows that Stephanie2 can proactively stop speaking or end the conversation at appropriate moments, such as when both sides say good night. Cases 2 and 3 further show that Stephanie1’s response delay varies drastically with output length, whereas Stephanie2 introduces a “thinking delay,” making the delay behavior more reasonable.
\section{Conclusion}
This paper targets the demand for step-by-step, multi-message interaction in instant-messaging social chat. We identify key limitations of existing step-by-step systems in active waiting and message pacing modeling, and propose Stephanie2, a step-wise decision-making dialogue agent. At each step, Stephanie2 explicitly decides whether to send or wait, and models latency as the sum of thinking time and typing time, making step-by-step conversations more consistent with human chatting patterns. We further introduce a time-window-based dual-agent dialogue framework to generate more natural step-by-step dialogue histories, supporting data construction and systematic evaluation. Experimental results show that Stephanie2 outperforms Stephanie1 in user-experience metrics such as naturalness and engagement, as well as in the role identification test. Our approach may benefit AI companion and emotional support applications.\\\textbf{Future work.} We plan to further explore multi-agent group chats under the step-by-step dialogue paradigm, and work toward stronger user-centered alignment and safety.

\section*{Limitations}

Due to resource constraints, we were not able to experiment on a broader set of larger-scale models, nor have we been able to allocate additional human resources for evaluation. Nevertheless, within the available budget, we carried out rigorous human evaluation and analysis with our best honest efforts, showing the effectiveness of Stephanie2. We look forward to further validating this step-by-step dialogue technique and assessing its effectiveness at scale in real-world consumer products.

\section*{Ethics Statement}

We honour and support the ACL Code of Ethics.
The datasets used in this work are well-known and
widely used, and the dataset pre-processing does
not make use of any external textual resource. In
our view, there is no known ethical issue. End-to-end pre-trained generators are also used, which
are subjected to generating offensive context. However,
the above-mentioned issues are widely known to
commonly exist for these models. Any content
generated does not reflect the view of the authors.



\bibliography{custom}

@article{zhang1801personalizing,
  title={Personalizing dialogue agents: I have a dog, do you have pets too? arXiv 2018},
  author={Zhang, Saizheng and Dinan, Emily and Urbanek, Jack and Szlam, Arthur and Kiela, Douwe and Weston, Jason},
  journal={arXiv preprint arXiv:1801.07243},
  year={2018}
}

@inproceedings{yang2025stephanie,
  title={Stephanie: Step-by-step dialogues for mimicking human interactions in social conversations},
  author={Yang, Hao and Lu, Hongyuan and Zeng, Xinhua and Liu, Yang and Zhang, Xiang and Yang, Haoran and Zhang, Yumeng and Huang, Shan and Wei, Yiran and Lam, Wai},
  booktitle={Findings of the Association for Computational Linguistics: NAACL 2025},
  pages={153--166},
  year={2025}
}

@inproceedings{wu2025x,
  title={X-TURING: Towards an Enhanced and Efficient Turing Test for Long-Term Dialogue Agents},
  author={Wu, Weiqi and Wu, Hongqiu and Zhao, Hai},
  booktitle={Proceedings of the 63rd Annual Meeting of the Association for Computational Linguistics (Volume 1: Long Papers)},
  pages={5874--5889},
  year={2025}
}

@article{gao2023retrieval,
  title={Retrieval-augmented generation for large language models: A survey},
  author={Gao, Yunfan and Xiong, Yun and Gao, Xinyu and Jia, Kangxiang and Pan, Jinliu and Bi, Yuxi and Dai, Yixin and Sun, Jiawei and Wang, Haofen and Wang, Haofen},
  journal={arXiv preprint arXiv:2312.10997},
  volume={2},
  number={1},
  year={2023}
}

@article{peng2024graph,
  title={Graph retrieval-augmented generation: A survey},
  author={Peng, Boci and Zhu, Yun and Liu, Yongchao and Bo, Xiaohe and Shi, Haizhou and Hong, Chuntao and Zhang, Yan and Tang, Siliang},
  journal={arXiv preprint arXiv:2408.08921},
  year={2024}
}

@article{schick2023toolformer,
  title={Toolformer: Language models can teach themselves to use tools},
  author={Schick, Timo and Dwivedi-Yu, Jane and Dess{\`\i}, Roberto and Raileanu, Roberta and Lomeli, Maria and Hambro, Eric and Zettlemoyer, Luke and Cancedda, Nicola and Scialom, Thomas},
  journal={Advances in Neural Information Processing Systems},
  volume={36},
  pages={68539--68551},
  year={2023}
}

@inproceedings{yao2022react,
  title={React: Synergizing reasoning and acting in language models},
  author={Yao, Shunyu and Zhao, Jeffrey and Yu, Dian and Du, Nan and Shafran, Izhak and Narasimhan, Karthik R and Cao, Yuan},
  booktitle={The eleventh international conference on learning representations}
}

@inproceedings{li2025review,
  title={A review of prominent paradigms for llm-based agents: Tool use, planning (including rag), and feedback learning},
  author={Li, Xinzhe},
  booktitle={Proceedings of the 31st International Conference on Computational Linguistics},
  pages={9760--9779},
  year={2025}
}

@article{luo2025large,
  title={Large language model agent: A survey on methodology, applications and challenges},
  author={Luo, Junyu and Zhang, Weizhi and Yuan, Ye and Zhao, Yusheng and Yang, Junwei and Gu, Yiyang and Wu, Bohan and Chen, Binqi and Qiao, Ziyue and Long, Qingqing and others},
  journal={arXiv preprint arXiv:2503.21460},
  year={2025}
}

@article{zhu2025graph,
  title={Graph-based Approaches and Functionalities in Retrieval-Augmented Generation: A Comprehensive Survey},
  author={Zhu, Zulun and Huang, Tiancheng and Wang, Kai and Ye, Junda and Chen, Xinghe and Luo, Siqiang},
  journal={arXiv preprint arXiv:2504.10499},
  year={2025}
}

@article{oche2025systematic,
  title={A systematic review of key retrieval-augmented generation (rag) systems: Progress, gaps, and future directions},
  author={Oche, Agada Joseph and Folashade, Ademola Glory and Ghosal, Tirthankar and Biswas, Arpan},
  journal={arXiv preprint arXiv:2507.18910},
  year={2025}
}

@inproceedings{kang2024can,
  title={Can large language models be good emotional supporter? mitigating preference bias on emotional support conversation},
  author={Kang, Dongjin and Mac Kim, Sunghwan and Kwon, Taeyoon and Moon, Seungjun and Cho, Hyunsouk and Yu, Youngjae and Lee, Dongha and Yeo, Jinyoung},
  booktitle={Proceedings of the 62nd Annual Meeting of the Association for Computational Linguistics (Volume 1: Long Papers)},
  pages={15232--15261},
  year={2024}
}

@inproceedings{zhi2024guidedempathy,
  title={GuidedEmpathy: Guiding Large Language Models for Empathetic Response Generation with Situational Awareness},
  author={Zhi, Longrun},
  booktitle={Proceedings of the 2024 8th International Conference on Computer Science and Artificial Intelligence},
  pages={354--360},
  year={2024}
}

@inproceedings{cao2024improving,
  title={Improving emotional support conversation with strategy-intent inference},
  author={Cao, Yaru and Yu, Hongzhi and Wan, Fucheng},
  booktitle={Proceedings of the 2024 8th International Conference on Electronic Information Technology and Computer Engineering},
  pages={227--233},
  year={2024}
}

@article{zhang2025intentionesc,
  title={IntentionESC: An Intention-Centered Framework for Enhancing Emotional Support in Dialogue Systems},
  author={Zhang, Xinjie and Wang, Wenxuan and Jin, Qin},
  journal={arXiv preprint arXiv:2506.05947},
  year={2025}
}

@inproceedings{zheng2023augesc,
  title={Augesc: Dialogue augmentation with large language models for emotional support conversation},
  author={Zheng, Chujie and Sabour, Sahand and Wen, Jiaxin and Zhang, Zheng and Huang, Minlie},
  booktitle={Findings of the Association for Computational Linguistics: ACL 2023},
  pages={1552--1568},
  year={2023}
}

@article{bai2025emotional,
  title={Emotional Supporters often Use Multiple Strategies in a Single Turn},
  author={Bai, Xin and Chen, Guanyi and He, Tingting and Zhou, Chenlian and Liu, Yu},
  journal={arXiv preprint arXiv:2505.15316},
  year={2025}
}

@inproceedings{wu2025personas,
  title={From personas to talks: Revisiting the impact of personas on llm-synthesized emotional support conversations},
  author={Wu, Shenghan and Zhu, Yimo and Hsu, Wynne and Lee, Mong-Li and Deng, Yang},
  booktitle={Proceedings of the 2025 Conference on Empirical Methods in Natural Language Processing},
  pages={5439--5453},
  year={2025}
}

@inproceedings{hao2025enhancing,
  title={Enhancing Emotional Support Conversations: A Framework for Dynamic Knowledge Filtering and Persona Extraction},
  author={Hao, Jiawang and Kong, Fang},
  booktitle={Proceedings of the 31st International Conference on Computational Linguistics},
  pages={3193--3202},
  year={2025}
}

@inproceedings{hashimoto2025career,
  title={A career interview dialogue system using large language model-based dynamic slot generation},
  author={Hashimoto, Ekai and Nakano, Mikio and Sakurai, Takayoshi and Shiramatsu, Shun and Komazaki, Toshitake and Tsuchiya, Shiho},
  booktitle={Proceedings of the 31st International Conference on Computational Linguistics},
  pages={1562--1584},
  year={2025}
}

@article{deng2025proactive,
  title={Proactive conversational ai: A comprehensive survey of advancements and opportunities},
  author={Deng, Yang and Liao, Lizi and Lei, Wenqiang and Yang, Grace Hui and Lam, Wai and Chua, Tat-Seng},
  journal={ACM Transactions on Information Systems},
  volume={43},
  number={3},
  pages={1--45},
  year={2025},
  publisher={ACM New York, NY}
}

@inproceedings{zhan2023socialdial,
  title={Socialdial: A benchmark for socially-aware dialogue systems},
  author={Zhan, Haolan and Li, Zhuang and Wang, Yufei and Luo, Linhao and Feng, Tao and Kang, Xiaoxi and Hua, Yuncheng and Qu, Lizhen and Soon, Lay-Ki and Sharma, Suraj and others},
  booktitle={Proceedings of the 46th International ACM SIGIR Conference on Research and Development in Information Retrieval},
  pages={2712--2722},
  year={2023}
}

@inproceedings{li2016diversity,
  title={A diversity-promoting objective function for neural conversation models},
  author={Li, Jiwei and Galley, Michel and Brockett, Chris and Gao, Jianfeng and Dolan, William B},
  booktitle={Proceedings of the 2016 conference of the North American chapter of the association for computational linguistics: human language technologies},
  pages={110--119},
  year={2016}
}

@article{bravo2025systematic,
  title={A Systematic Review on Artificial Intelligence-Based Multimodal Dialogue Systems Capable of Emotion Recognition},
  author={Bravo, Luis and Rodriguez, Ciro and Hidalgo, Pedro and Angulo, Cesar},
  journal={Multimodal Technologies and Interaction},
  volume={9},
  number={3},
  pages={28},
  year={2025},
  publisher={MDPI}
}

@article{chen2025future,
  title={The future of cognitive strategy-enhanced persuasive dialogue agents: new perspectives and trends},
  author={Chen, Mengqi and Guo, Bin and Wang, Hao and Li, Haoyu and Zhao, Qian and Liu, Jingqi and Ding, Yasan and Pan, Yan and Yu, Zhiwen},
  journal={Frontiers of Computer Science},
  volume={19},
  number={5},
  pages={195315},
  year={2025},
  publisher={Springer}
}

@article{guo2025seed1,
  title={Seed1. 5-vl technical report},
  author={Guo, Dong and Wu, Faming and Zhu, Feida and Leng, Fuxing and Shi, Guang and Chen, Haobin and Fan, Haoqi and Wang, Jian and Jiang, Jianyu and Wang, Jiawei and others},
  journal={arXiv preprint arXiv:2505.07062},
  year={2025}
}

@article{algherairy2024review,
  title={A review of dialogue systems: current trends and future directions},
  author={Algherairy, Atheer and Ahmed, Moataz},
  journal={Neural Computing and Applications},
  volume={36},
  number={12},
  pages={6325--6351},
  year={2024},
  publisher={Springer}
}

@article{liu2024deepseek,
  title={Deepseek-v3 technical report},
  author={Liu, Aixin and Feng, Bei and Xue, Bing and Wang, Bingxuan and Wu, Bochao and Lu, Chengda and Zhao, Chenggang and Deng, Chengqi and Zhang, Chenyu and Ruan, Chong and others},
  journal={arXiv preprint arXiv:2412.19437},
  year={2024}
}

@article{dubey2024llama,
  title={The llama 3 herd of models},
  author={Dubey, Abhimanyu and Jauhri, Abhinav and Pandey, Abhinav and Kadian, Abhishek and Al-Dahle, Ahmad and Letman, Aiesha and Mathur, Akhil and Schelten, Alan and Yang, Amy and Fan, Angela and others},
  journal={arXiv preprint arXiv:2407.21783},
  year={2024}
}

\appendix

\section{Case Study}
\label{sec:appendix}
\begin{figure*}[h!t]
\centering
\includegraphics[width=\textwidth]{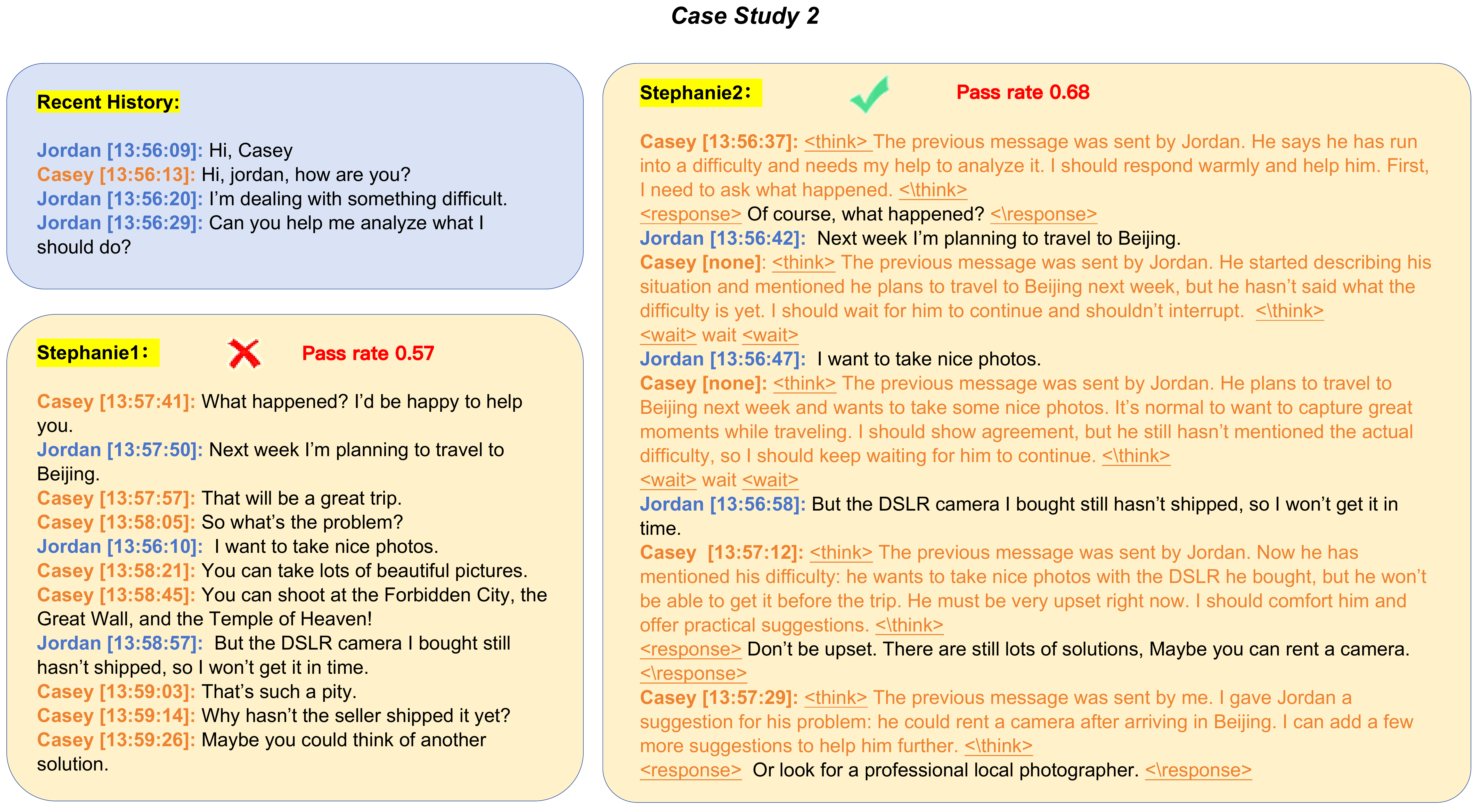}
\caption{Case 2: Stephanie can keep waiting and listening if the other person has not yet finished providing additional details.}
\label{fig: case2}
\end{figure*}

\begin{figure*}[h!t]
\centering
\includegraphics[width=\textwidth]{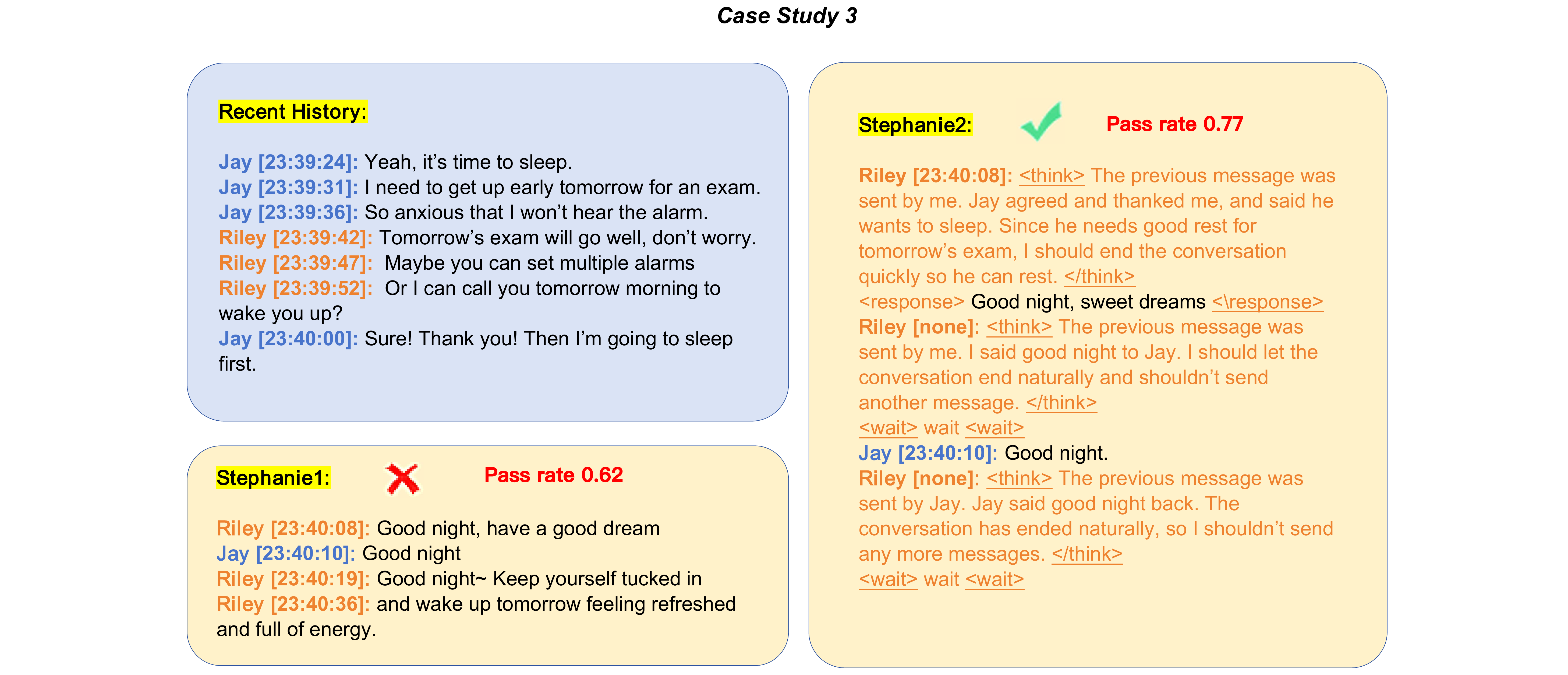}
\caption{Case 3: Stephanie can proactively end the conversation, for example, when both sides say good night.}
\label{fig: case3}
\end{figure*}

Cases 1–3 are shown in Figs.~\ref{fig: method}, \ref{fig: case2}, and \ref{fig: case3}, respectively.

\section{Topic Distribution}\label{topic_distribution}

\begin{figure}[h]
\centering
\includegraphics[width=0.48\textwidth]{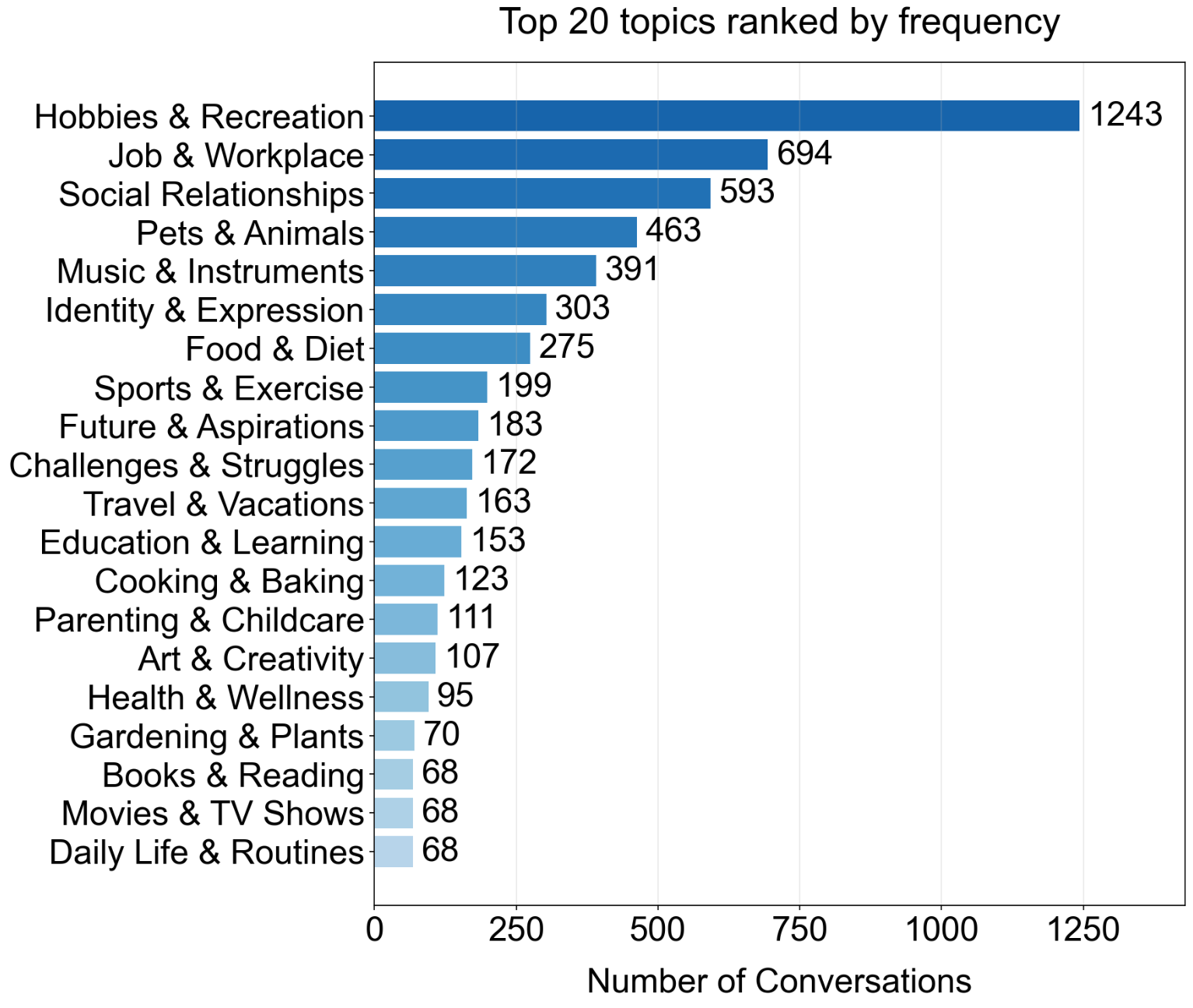}
\caption{Topic distribution.}
\label{fig: topic distribution}
\end{figure}

The distribution of the top 20 topics is shown in Fig.~\ref{fig: topic distribution}.

\section{Prompt}\label{prompt}

\begin{tcolorbox}[breakable, colback=white, colframe=black, title=Stephanie2 Agent Prompt]
\begin{Verbatim}[
  fontsize=\small,
  breaklines=true,
  breakanywhere=true,
  breaksymbolleft={},
  breaksymbolright={}
]
Chat History:
<|HISTORY|>

Task:
You are <|NAME2|>, and your persona is "<|PERSONALITY2|>". You are chatting with your friend <|NAME1|> on WeChat about "<|TOPIC|>". The above is your chat history. Please think in the following steps:
"Who sent the last message in the chat history? First, restate the content of the last message. If the last message was sent by the other person, think about whether you should respond or wait for them to continue. If the last message was sent by yourself, you need to think and decide whether to wait for the other person to reply or continue sending a message. If you choose to send a message, think about what to say next."

If you think you should send a message, you may draft multiple short messages. Output in the following format:
<think> Write your thoughts here. <\think>
<response> Only write the next single short message you want to send here, based on your thoughts. <\response>

If you want to wait, output in the following format:
<think> Write your thoughts here. <\think>
<wait> wait <\wait>
\end{Verbatim}
\end{tcolorbox}

\begin{tcolorbox}[breakable, colback=white, colframe=black, title=Prompt for the Role Identification Test]
\begin{Verbatim}[
  fontsize=\small,
  breaklines=true,
  breakanywhere=true,
  breaksymbolleft={},
  breaksymbolright={}
]
You are doing a variant of Turing test. Below is a conversation between two people, one is a real human and one is an AI. Please determine which role is the AI.

Conversation:
{dialogue}

Please answer with only one of the following three options:
A. Role 1 is AI  
B. Role 2 is AI
C. Hard to tell

Wrap your final answer in <answer></answer> tags.
\end{Verbatim}
\end{tcolorbox}

\begin{tcolorbox}[breakable, colback=white, colframe=black, title=Prompt for Long-term Memory Summarization]
\begin{Verbatim}[
  fontsize=\small,
  breaklines=true,
  breakanywhere=true,
  breaksymbolleft={},
  breaksymbolright={}
]
You are a dialogue summarization assistant. Please merge the previous dialogue summary and the recent conversation records to generate a new summary. Be sure to retain important information such as key times, locations, people, events, and emotional changes. Output only the summary content directly, without adding any prefix.

Previous dialogue summary:
<|EXISTING_SUMMARY|>

Recent conversation records:
<|CONVERSATIONS|>
\end{Verbatim}
\end{tcolorbox}

\begin{tcolorbox}[breakable, colback=white, colframe=black, title=Prompt for Step-by-Step Rewriting of Persona-Chat]
\begin{Verbatim}[
  fontsize=\small,
  breaklines=true,
  breakanywhere=true,
  breaksymbolleft={},
  breaksymbolright={}
]
You are a data conversion assistant. Your task is to convert PersonaChat dataset entries into a conversation format.

Given:
1. Two personas (your persona and partner's persona)
2. A conversation history between them

Please output a JSON object with the following structure:
{
  "topic": "<summarize the main conversation topic in ~10 words or less>",
  "characters": [
    {
      "name1": "<generate a short name for persona 1>",
      "personality": "<keep the original persona descriptions as a list, joined by '. '>"
    },
    {
      "name2": "<generate a short name for persona 2>",
      "personality": "<keep the original persona descriptions as a list, joined by '. '>"
    }
  ],
  "recent_conversations": [
    {
      "timestamp": "<generate realistic timestamp>",
      "role": "<name1 or name2>",
      "content": "<message content>"
    },
    ...
  ]
}

Rules:
1. Generate simple names like "Alex", "Sam", "Jordan", "Riley", "Casey" etc
2. Keep the original persona descriptions for personality
3. People can send multiple short messages
4. Generate realistic timestamps between messages
5. Add a "topic" field summarizing the conversation theme in 10 words or less
6. Output ONLY the JSON, no other text

Here are 5 examples:

<|EXAMPLE1|>
<|EXAMPLE2|>
<|EXAMPLE3|>
<|EXAMPLE4|>
<|EXAMPLE5|>

Now convert the following input:

Input format:
<|PERSONAS|>
<|CONVERSATION|>
\end{Verbatim}
\end{tcolorbox}

\begin{tcolorbox}[breakable, colback=white, colframe=black, title=First-level Prompt for Hierarchical Topic Summarization]
\begin{Verbatim}[
  fontsize=\small,
  breaklines=true,
  breakanywhere=true,
  breaksymbolleft={},
  breaksymbolright={}
]
You are a topic clustering assistant. Given the following 600 conversation topics, 
identify 60 broader topic categories that can cover all of them.

Topics:
{topics_json}

Output a JSON array with exactly 60 topic names (strings only, 5-10 words each).

Example output:
["Hobbies and leisure activities", "Work and career", "Family and relationships", ...]

Output ONLY the JSON array of topic names, no other text.
\end{Verbatim}
\end{tcolorbox}

\begin{tcolorbox}[breakable, colback=white, colframe=black, title=Second-level Prompt for Hierarchical Topic Summarization]
\begin{Verbatim}[
  fontsize=\small,
  breaklines=true,
  breakanywhere=true,
  breaksymbolleft={},
  breaksymbolright={}
]
You are a topic clustering assistant. Given the following 660 topic categories,
consolidate them into exactly 60 final broader topic categories.

Topics:
{topics_json}

Output a JSON array with exactly 60 topic names (strings only, 5-10 words each).
Try to make the categories balanced and cover all the original topics.

Example output:
["Hobbies and leisure activities", "Work and career", "Family and relationships", ...]

Output ONLY the JSON array of topic names, no other text.
\end{Verbatim}
\end{tcolorbox}

\begin{tcolorbox}[breakable, colback=white, colframe=black, title=The Prompt for Assigning A Topic to Each Dialogue]
\begin{Verbatim}[
  fontsize=\small,
  breaklines=true,
  breakanywhere=true,
  breaksymbolleft={},
  breaksymbolright={}
]
You are a topic clustering assistant. Assign each of the following conversations to ONE of the given topics.

Available topics:
{topics_list}

Conversations to assign:
{topics_json}

Output a JSON object where keys are topic names and values are arrays of IDs assigned to that topic.
Each ID must appear exactly once. Use only the topic names provided above.

Example output:
{{
  "Hobbies and leisure activities": [1, 5],
  "Work and career": [2, 3, 4]
}}

Output ONLY the JSON object, no other text.
\end{Verbatim}
\end{tcolorbox}

\end{document}